\documentclass{article}
\usepackage[margin=3.0cm]{geometry}

\usepackage{graphicx} 

\usepackage{amsfonts,amsmath,amssymb,amsthm,mathtools}
\usepackage{relsize}
\usepackage{hyperref}
\usepackage{tikz}
\usepackage{xcolor}

\newtheorem{theorem}{Theorem}[section]

\newtheorem{proposition}[theorem]{Proposition}

\newtheorem*{corollary*}{Corollary}

\theoremstyle{definition}
\newtheorem{definition}[theorem]{Definition}
\newtheorem{example}[theorem]{Example}

\theoremstyle{remark}

\title{Persistent reachability homology in machine learning applications}
\author{Luigi Caputi, Nicholas Meadows, Henri Riihimäki}

\begin{document}

\maketitle

\begin{abstract}
    We explore the recently introduced persistent reachability homology (PRH) of digraph data, i.e.~data in the form of directed graphs. In particular, we study the effectiveness of PRH in network classification task in a key neuroscience problem: epilepsy detection. PRH is a variation of the persistent homology of digraphs, more traditionally based on the directed flag complex~(DPH). A main advantage of PRH is that it considers the condensations of the digraphs appearing in the persistent filtration and thus is computed from smaller digraphs. We compare the effectiveness of PRH to that of DPH and we show that PRH outperforms DPH in the classification task. We use the Betti curves and their integrals as topological features and implement our pipeline on support vector machine.
\end{abstract}

\section*{Introduction}
\emph{Topological Data Analysis} (TDA) is a fast-growing research field at the intersection of algebraic topology, data analysis, computational geometry and topology, machine learning and statistics. The topological approach to data analysis was initiated with the works~\cite{frosini_1990,elz,ez,carlssondata}, and since then interest in the field skyrocketed. Nowadays, we are witnessing to applications of TDA in various fields, such as neuroscience, chemistry, finance, material science, and image classification, to name a few. Persistent homology (PH) is one of the main tools adopted in TDA, and it is a multi-scale adaptation of the classical (simplicial) homology theories. It is readily computable, and it was shown that PH is stable with respect to small noise perturbations of the data~\cite{stabilityph}, which makes it a reliable analytics tool.

The main ingredient in employing TDA tools, and persistence methods in particular, in machine learning pipelines is the homology theory used for extracting homological features from data. For point cloud data, i.e.\ finite point sets embedded in a metric space, 
simplicial homology is the most common choice, 
both for its efficient computability and geometric interpretation. Data in the form of undirected graphs can equally well resort to simplicial homology via the clique (also called flag) complex construction, where any clique of \(k+1\) vertices spans a \(k\)-simplex in the clique complex. A point cloud can even be encoded into an edge-weighted graph where the edge weights are determined by the
metric. 

One of the prominent applications of TDA and PH the authors are interested in is in neuroimaging and neuroscience; see, e.g.~\cite{8052510,8352585,lee2019} for some contributions and reviews in this subject. Developing sensitive and reliable methods to distinguish normal and abnormal brain states is in fact a key neuroscientific challenge, and the study of complex patterns of brain network topology has become a flourishing area of research~\cite{Bullmore2009}. In~\cite{Caputi2021} it was shown that specific applications for TDA may arise when a direct comparison of connectivity matrices is not suitable, such as for intracranial electrophysiology with an individual number and location of measurements for each subject. A related open problem is to determine the most relevant persistent homology features when the data are represented in the form of digraphs (directed graphs) -- such as for Granger causality and information networks. With the aim of developing new tools capable of capturing significant topological features of such data and to test their effectiveness, in this work we shall focus on a main neuroscientific classification task: epilepsy detection from EEG correlation networks. Consequently, we shall focus on digraph data and explore a novel persistent homology theory in machine learning applications: \emph{persistent reachability homology} (PRH)~\cite{zbMATH07844814}. 

Digraph data is prevalent, arising from synaptic neuronal connections, citations and followings in scientific and social contexts, protein interactions, and web hyperlinks, to name a few examples. Digraphs can also be endowed with a simplicial structure via the directed flag complex, where \(k\)-simplices are spanned by directed \(k+1\)-cliques. This facilitates the computation of simplicial (persistent) homology of digraphs, with an efficient implementation~\cite{Flagser_paper}. Furthermore, this approach has already been successfully applied in network neuroscience \cite{Frontiers_paper}, in constructing a machine learning pipeline for classifying network dynamics \cite{Tribes_math, Tribes_neuroscience}, and exploring the structure and topology of biological and artificial neural networks \cite{  DNN_homologies, Govc_2020,Caputi2021,Riihimaki_simplicial_connectivities,Carannante_simplicial}.
However, the homology of the directed flag complex might fail to differentiate between very different networks because of its homotopy invariance. The illustration below gives a concrete demonstration of this effect; both digraphs have the topology of a circle, even though the left digraph exhibits a much more intricate network structure.

\begin{center}
	\begin{tikzpicture}
		\tikzstyle{point}=[circle,thick,draw=black,fill=black,inner sep=0pt,minimum width=2pt,minimum height=2pt]
		\tikzstyle{arc}=[thick,shorten >= 4pt,shorten <= 4pt,->]
		\tikzstyle{arc2}=[thick,shorten >= 9pt,shorten <= 9pt,->]
		
		\coordinate (a) at (-0.5,1);
		\coordinate (b) at (1,1.85);
		\coordinate (c) at (0.75,-0.4);
		\coordinate (d) at (-2,-1.25);
		\coordinate (e) at (2.25,-0.5);
		\coordinate (f) at (3,1.5);
		\coordinate (g) at (-2.2,1.5);
		
		\draw[white,line width=5pt,fill=blue,fill opacity=0.2] (d) -- (a) -- (c) --cycle;
		\draw[white,line width=5pt,fill=blue,fill opacity=0.2] (d) -- (g) -- (a) --cycle;
		\draw[white,line width=5pt,fill=blue,fill opacity=0.2] (a) -- (g) -- (b) --cycle;		
		\draw[white,line width=5pt,fill=blue,fill opacity=0.2] (e) -- (c) -- (b) --cycle;		
		\draw[white,line width=5pt,fill=blue,fill opacity=0.2] (f) -- (e) -- (b) --cycle;
		
		\draw[arc] (b) -- (a);
		\draw[arc] (a) -- (c);
		\draw[arc] (d) -- (a);
		\draw[arc] (d) -- (c);
		\draw[arc] (c) -- (b);
		\draw[arc] (b) -- (e);
		\draw[arc] (c) -- (e);
		\draw[arc] (b) -- (f);
		\draw[arc] (e) -- (f);
		\draw[arc] (d) -- (g);
		\draw[arc] (g) -- (a);
		\draw[arc] (b) -- (g);
		
		\node[point] at (a) {};
		\node[point] at (d) {};
		\node[point] at (c) {};
		\node[point] at (b) {};
		\node[point] at (e) {};
		\node[point] at (f) {};
		\node[point] at (g) {};
		
		
		\node[] at (3.35,0.35) {\(\approx\)};
		
		\coordinate (a') at (4.55,1);
		\coordinate (c') at (5.8,0.35);
		\coordinate (d') at (3.8,-0.5);
		
		\draw[arc] (a') -- (c');
		\draw[arc] (d') -- (a');
		\draw[arc] (c') -- (d');
		
		\node[point] at (a') {};
		\node[point] at (d') {};
		\node[point] at (c') {};
		
	\end{tikzpicture}
\end{center}

This type of ignorance raises the fundamental question if there is a homology theory more sensitive to the digraph structure. Such a homology theory could form a novel basis for persistence and ensuing machine learning applications for network data. One possible approach comes from Hochschild (co)homology \(\mathrm{HH}\), which is a (co)homology theory of $k$-algebras~\cite{loday} for \(k\) a coefficient field. In \cite{persistentHH}, a persistent Hochschild (co)homology pipeline was constructed, with applications to network analysis. Every digraph~\(G\), in fact, gives rise to the so-called \emph{path algebra}, i.e.~the algebra~\(kG\) generated by all the (directed) paths in \(G\), with product given by concatenation of paths; hence, we can consider the associated (co)homology \(\mathrm{HH}(kG)\). It turns out that for acyclic digraphs, i.e.\ feedforward type networks, the cohomology Betti numbers of \(\mathrm{HH}(kG)\) can be computed with an explicit combinatorial formula. Moreover, to give a positive answer to the above question concerning homological sensitivity to digraph structure, it is easy to construct examples of digraphs whose directed flag complexes are topologically trivial, while the cohomology \(\mathrm{HH}(kG)\) has nontrivial Betti numbers, as the illustrations below show: 

\begin{center}
	\begin{tikzpicture}
		\tikzstyle{point}=[circle,thick,draw=black,fill=black,inner sep=0pt,minimum width=2pt,minimum height=2pt]
		\tikzstyle{arc}=[thick,shorten >= 4pt,shorten <= 4pt,->]
		
		\coordinate (a) at (0,0);
		\coordinate (b) at (0.75,1);
		\coordinate (c) at (1.5,0);
		
		\draw[arc] (a) -- (b);
		\draw[arc] (b) -- (c);
		\draw[arc] (a) -- (c);
		
		\node[point] at (a) {};
		\node[point] at (b) {};
		\node[point] at (c) {};

        \node[] at (0.75,-0.7) {\(\beta^1_{HH}=2\)};


		\coordinate (a1) at (2.5,0.5);
		\coordinate (b1) at (3.4,1.2);
		\coordinate (c1) at (3.4,-0.2);
        \coordinate (d1) at (4.3,0.5);
		
		\draw[arc] (a1) -- (b1);
		\draw[arc] (b1) -- (c1);
		\draw[arc] (a1) -- (c1);
        \draw[arc] (b1) -- (d1);
        \draw[arc] (c1) -- (d1);
		
		\node[point] at (a1) {};
		\node[point] at (b1) {};
		\node[point] at (c1) {};
        \node[point] at (d1) {};

        \node[] at (3.4,-0.7) {\(\beta^1_{HH}=4\)};


		\coordinate (a2) at (5.3,0.5);
		\coordinate (b2) at (6.2,1.2);
		\coordinate (c2) at (6.2,-0.2);
        \coordinate (d2) at (7.1,0.5);
		
		\draw[arc] (a2) -- (b2);
		\draw[arc] (b2) -- (c2);
		\draw[arc] (a2) -- (c2);
        \draw[arc] (b2) -- (d2);
        \draw[arc] (d2) -- (c2);
		
		\node[point] at (a2) {};
		\node[point] at (b2) {};
		\node[point] at (c2) {};
        \node[point] at (d2) {};

        \node[] at (6.2,-0.7) {\(\beta^1_{HH}=5\)};


		\coordinate (a3) at (8.1,0.5);
		\coordinate (b3) at (9,1.2);
		\coordinate (c3) at (9,-0.2);
        \coordinate (d3) at (9.9,0.5);
		
		\draw[arc] (a3) -- (b3);
		\draw[arc] (b3) -- (c3);
		\draw[arc] (a3) -- (c3);
        \draw[arc] (b3) -- (d3);
        \draw[arc] (c3) -- (d3);
        \draw[arc] (a3) -- (d3);
		
		\node[point] at (a3) {};
		\node[point] at (b3) {};
		\node[point] at (c3) {};
        \node[point] at (d3) {};

        \node[] at (9,-0.7) {\(\beta^1_{HH}=8\)};
	\end{tikzpicture}
\end{center}

The development of new homology theories to be used in persistent homology pipelines is then fundamental in order to capture information on network data that is complementary to, and sometimes ignored by classical simplicial methods. As a consequence, TDA based machine learning can benefit from different types of homological information having different expressive powers. The directed flag complex captures the global topological organisation of the directed cliques, each of which can be regarded as a small feedforward computational unit within the network. Hochschild (co)homology of the path algebra in turn captures information about the combinatorics in the organisation of all paths in the network. As shown in the above examples, the path algebra, in conjunction with Hochschild cohomology computations, can provide finer homological information which might translate into more expressive feature vectors in network classification tasks.

The reachability construction considered in this work was introduced in \cite{zbMATH07844814}. It fixes the non-functoriality in the persistent Hoschschild (co)homology pipeline of \cite{persistentHH} which results from the utilisation of the condensation operation of digraphs in order to make use of the formula in Theorem \ref{thmhapp} for efficient computation of Betti numbers and persistent Betti curves. The reachability digraph $\mathrm{Reach}(G)$ of \(G\) is essentially the transitive closure of \(G\). However, the homological information in $\mathrm{Reach}(G)$ is that of the poset obtained by condensing all the strongly connected components of \(G\) into single vertices. Hence, as compared to the information in the path algebra, $\mathrm{Reach}(G)$ captures only the existence of some path between vertices, and in particular the existence of non-recurring paths. This can in some cases be beneficial in reducing noise and simplifying the network data at hand. Furthermore, reachability homology recently appeared in the context of magnitude homology~\cite{hepworth2023reachability}, and its decategorification -- the so-called \emph{magnitude} -- was source of investigations in TDA in~\cite{NEURIPS2024_dfc24bd3}.

In this paper we undertake the initial investigation into the application of reachability homology for a network classification task, and we compare its performance with that of the homology of the directed flag complex. In Section \ref{sec:background} we give the mathematical background on the directed flag complex and its homology, reachability homology, and the relation of the latter to Hochschild (co)homology. In Section \ref{sec:random_graphs} we illustrate the expected low degree Betti numbers of reachability digraphs and directed flag complexes on a sample of Erdõs-Rényi random digraphs \(G(n,p)\) on \(n=100\) vertices for a range of edge probability values \(p\). These computations provide background to choices taken in our network classification pipeline.

The steps of the classification pipeline are outlined in Section \ref{sec:pipeline}. We employ feature vectors constructed from the Betti curves of reachability and directed flag complex homology, as well as the integrals of the respective Betti curves. As a classification task we perform epileptic seizure detection from EEG correlation networks of various subjects, using all the mentioned featurisation methods. Our pipeline is implemented using a support vector machine with both linear and RBF kernel and all classification results are presented in Section \ref{subsec:epilepsy_results}. 

Our results show that reachability homology yields the highest classification accuracy in all but one of the 8 different model comparisons explained in Section \ref{subsec:seizure_detection}. The top accuracy we observe in our experiments is 82\%. The linear SVM kernel allows to extract feature importance via feature ranking. We present these results also in Section~\ref{subsec:epilepsy_results} in the case of Betti curve features, as these have more direct topological interpretation. We observe that directed flag complex and reachability use different topological information. In particular, homology of the directed flag complex relies heavily on the Betti number in degree~2. This is an interesting independent observation, since in TDA applications there is a tendency to focus on using Betti numbers 0 and 1, largely due to computational efficiency. 







\section{Background}\label{sec:background}

In the following, by a graph we always mean a \emph{finite} (directed) simple graph; that is, we do not allow multiple edges. 
By a weighted (directed) graph we shall mean a (directed) graph with real-valued labellings of the edges. All homology and cohomology groups throughout the paper are taken with field coefficients, unless otherwise specified. 

For a weighted graph $G$ with $N$ edges, it is customary in topological data analysis to compute persistent homology invariants from the sequence of clique complexes, also known as ``flag complexes'' associated to $G$. For completeness, we briefly recall the construction. 

We start by filtering an undirected graph $G$ by thresholding the weights on the edges: if $w_0 < \dots < w_N$ are the ordered weights of the edges of $G$, we define $G[w_i]$ to be the induced (unweighted) subgraph of $G$ consisting of the same vertices as $G$, and with edges precisely the edges of $G$ of weight  $\leq w_i$. This yields a filtration of graphs
\begin{equation}\label{eq:graph_filtration}
G[w_{0}]\rightarrow G[w_{1}] \rightarrow \dots \rightarrow G[w_{N}] \ ,
\end{equation}
that is a sequence of undirected graphs and inclusions.
The clique complex of an undirected graph $H$ is the simplicial complex $\mathrm{Fl}(H)$ with simplices the complete subgraphs of $H$. Hence, for the weighted graph $G$, we obtain the filtration
\[
\mathrm{Fl}(G[w_{0}])\rightarrow \mathrm{Fl}(G[w_{1}]) \rightarrow \dots \rightarrow \mathrm{Fl}(G[w_{N}]) \ 
\]
of clique complexes. The persistent homology groups of $G$ are then defined as the persistent homology groups associated to the resulting filtration of clique complexes. 

In the case of $H$ a digraph (directed graph), we denote by $\mathrm{dFl}(H)$ the \emph{directed flag complex} associated to $H$. Before recalling its definition we recall that a directed $n$-clique of $H$ is a subgraph of $H$ on a collection of vertices $(v_1,\dots, v_n)$ with the property that there is a directed edge $(v_i,v_j)$ if and only if $i<j$. Then,  as in the undirected case, we define $\mathrm{dFl}(H)$ to be the  simplicial complex on the directed cliques of $H$. If $G$ is a directed weighted graph, we consider the filtration of directed graphs
\[
G[w_{0}]\rightarrow G[w_{1}] \rightarrow \dots \rightarrow G[w_{N}] 
\]
given by the weights, and the filtration 
\[
\mathrm{dFl}(G[w_{0}])\rightarrow \mathrm{dFl}(G[w_{1}]) \rightarrow \dots \rightarrow \mathrm{dFl}(G[w_{N}]) 
\]
of directed flag complexes. The persistent homology groups of the weighted directed graph~$G$ are then defined as the persistent homology groups associated to the filtration of directed flag complexes. 

\subsection{Reachability homology}

As reviewed in the previous section, the classical persistent homology groups associated to a (weighted) digraph are the persistent homology groups associated to the filtration of directed flag complexes. In this section we recall a different homology theory of digraphs called \emph{reachability homology}~\cite{zbMATH07844814,hepworth2023reachability}, which will take a prominent role in our theory. 

For a directed graph $G$, we consider the reachability digraph $\mathrm{Reach}(G)$. This is the directed graph on the same vertices of $G$ and with a directed edge $(v,w)$ from $v$ to $w$ if and only if there is a directed path from $v$ to $w$ in $G$. This \emph{reachability relation} endows $\mathrm{Reach}(G)$ with the structure of a preorder.

Recall that a digraph is \emph{strongly connected} if it contains a directed paths from $x$ to $y$ and from $y$ to~$x$, for every pair of vertices $x$ and $y$. A subgraph \(G' \subset G\) is a \emph{strongly connected component} of \(G\) if it is strongly connected and maximal with respect to this property. The \emph{condensation} \(c(G)\) of \(G\) is the digraph with the strongly connected components of~\(G\) as vertices; for two distinguished vertices~\(X\) and \(Y\) there is a directed edge \((X,Y)\) in \(c(Q)\) if and only if there is an edge \((x,y)\) in~\(G\) for some \(x \in X\) and \(y \in Y\).   Therefore, we can consider the condensation of $\mathrm{Reach}(G)$. This is a partially ordered set (a poset) with elements the strongly connected components of $G$ and with relation $X\leq Y $ if and only if there is a directed edge $X\to Y$ in $G$. The advantage of taking the condensation is that it kills all directed cycles in (the reachability digraphs of) $G$. We denote by $\mathcal{R}(G)$ the poset $c(\mathrm{Reach}(G))$.

\begin{definition}
    The poset $\mathcal{R}(G)$ is called the \emph{reachability poset} of $G$.
\end{definition}

Observe that, if $G$ is strongly connected, then its reachability poset consists of a single vertex. On the other hand, if $G$ does not contain directed cycles, then the reachability poset of $G$ is nothing but its transitive closure.

Recall that the homology of a poset $(P,\leq)$ is the homology of its order complex; the order complex of a poset $(P,\leq)$ is the simplicial complex on the vertex set $P$ whose $k$-simplices are the chains $x_0 < \cdots < x_k$ of $P$. We define the reachability homology of digraphs as follows:

\begin{definition}
The \emph{reachability homology} $\mathrm{RH}_*(G)$ of a digraph $G$ is the homology of the poset $\mathcal{R}(G)$.  
\end{definition}

\begin{example}
    Let $T_n$ be the transitive tournament on $n$ vertices; that is $T_n$ has vertices $\{1,\dots,n\}$ and directed edges $(i,j)$ for each $i<j$. Then, the reachability homology of $T_n$ is trivial; in fact we have
    \[
    \mathrm{RH}_*(G)=
    \begin{cases}
        \mathbb{Z} & \text{ if } *=0;\\
        0 & \text{ otherwise};
    \end{cases}
    \]
    where the reachability homology is taken with integer coefficients.
\end{example}

The reachability homology of transitive tournaments is always trivial because the posets $\mathcal{R}(G)$ have a minimal element. To get non-trivial homology, we have to consider digraphs without minimal elements. In the following example we see that this is not a sufficient condition:

\begin{example}
Let $G$ be a strongly connected directed graph. Then, the condensation of the transitive closure of $G$ is trivial, as it consists of a single vertex -- corresponding to the strongly connected component $G$ -- and a self loop. The reachability homology of $G$ is then trivial. More generally, let $G$ be an undirected connected graph, and let $\overline{G}$  be the directed graph associated to $G$; that is, $\overline{G}$  has same the vertices as $G$. If $\{v,w\}$ is an undirected edge of $G$, we add the directed edges $(v,w)$ and $(w,v)$ to  $\overline{G}$. As $G$ is connected, $\overline{G}$ is a strongly connected digraph, and the reachability homology of $\overline{G}$ is trivial.
\end{example}

We now provide examples of non-trivial reachability homology groups. First, recall that if $(P,\leq)$ is a poset, then we can consider the associated digraph $G(P)$  with vertices the elements of $P$ and with a directed edge $p\to q$ if and only if $p\leq q$ in $P$. 

\begin{example}\label{ex:reachabilityposets}
Let $K$  be a finite simplicial complex, and consider the face poset $P(K)$ of $K$; this is the poset consisting of the simplices of $K$ ordered by inclusion. Then, let $G(K)=G(P(K))$ be the associated digraph. As $P(K)$ is a poset, $G(K)$ is an acyclic digraph, and it is already transitively closed. Hence, $\mathcal{R}(G(K))=G(K)$. The reachability homology of $G(K)$ is the (standard) homology of the poset $P(K)$, hence we have
\[\mathrm{RH}_i(G(K))=\mathrm{H}_i(K)\]
for all $i\in \mathbb{N}$. 
\end{example}

In general, the advantage of computing reachability homology of digraphs, rather than the homology of the directed flag complex, is that the condensation drastically reduces the size of the digraphs, killing the strongly connected components and preserving the relevant connections. On the other hand, if the digraph is without directed cycles, for example, if $G$ is the digraph associated to a poset, then the reachability homology of $G$ and the homology of the directed flag complex are isomorphic: 

\begin{proposition}
        Let $P$ be a poset and $G(P)$ its underlying directed graph. Then, we have
    \[
    \mathrm{RH}_i(G(P))=\mathrm{H}_i(\mathrm{dFl}(G(P)))
    \]
        for all $i\in \mathbb{N}$. 
\end{proposition}

\subsection{Reachability homology and Hochschild cohomology}

Our interest in reachability homology arises from its relationship to an important homology theory of algebras, called \emph{Hochschild (co)homology}. In this section we clarify this relationship, starting with recalling the main definition of Hochschild homology, following \cite[Section~1.1]{loday}.

For a commutative ring $R$, let  $A$ be an associative unital $R$-algebra; for example, $A$ can be a polynomial algebra over~$R$. Let $C_n(A)$ be the $R$-module  $$C_n(A)\coloneqq A^{\otimes n+1} \ ,$$ defined as the tensor product of  $n+1$ copies of $A$ (all the tensor products being  over $R$). The boundary operator, classically denoted by $b$, is the $R$-linear map $b\colon  C_n(A)\to C_{n-1}(A)$ defined as follows:
\[
\begin{split}
  b(a_0,a_1,\dots,a_n)=  &(a_0 a_1,a_2,\dots,a_n)+\\ &+\sum_{i=1}^{n-1} (-1)^i(a_0,a_1,\dots, a_i a_{i+1},\dots,a_n)+\\&+ (-1)^n(a_n a_0,a_1,\dots, a_{n-1}) \ .
\end{split}
\]
In the formula, for simplicity of notation, we have dropped the tensor products.
The map $b$ is a boundary operator \cite[Lemma~1.1.2]{loday} and the pair $(C_*(A),b)$ is a chain complex, called the \emph{Hochschild complex}.

\begin{definition}\label{defhh}
	The \emph{Hochschild homology} groups $\mathrm{HH}_*(A)$ of an associative unital algebra $A$ (with coefficients in  $A$) are the homology groups of the Hochschild complex. Hochschild cohomology of $A$ is the homology of the dual complex.
\end{definition}

To each directed graph we can associate a standard algebra, called \emph{path algebra}. Let $k$ be a field, and denote by $s(e)$ and $t(e)$ the source and target of a directed edge \(e\) in $G$, respectively.

\begin{definition}\label{defpathalg}
	The \emph{path algebra} $kG$ associated to the digraph $G$ is the $k$-vector space with a basis consisting of all possible paths in $G$, and the multiplication being defined 
	on two basis paths \(\gamma = (e_1,\dots,e_n)\), \(\gamma' = (e'_1,\dots,e'_p)\) by the formula
	\[\gamma \gamma' = \begin{cases}
		(e_1,\dots,e_n,e'_1,\dots,e'_p), &\text{ if } s(e'_1)=t(e_n)\\
		0, &\text{ otherwise} 
	\end{cases} \ .\]
\end{definition}

The path algebra $kG$ associated to a digraph $G$ is an associative algebra over $k$, and has a unit if the digraph is finite -- see, eg.~\cite[Lemma~2.21]{persistentHH}. 

Computations of Hochschild (co)homology groups may be difficult for arbitrary associative algebras, but when $A$ is the path algebra~$kG$ of a directed graph, computations are easier and reflect the combinatorial properties of the digraph~$G$. First, it is a standard fact that the Hochschild cohomology groups  $\mathrm{HH}^*(A)$ of the path algebra $A$ vanish in degrees $\geq 2$.  
In degrees $0$ and $1$, the computation of Hochschild cohomology is due to Happel~\cite{happel} 
(see also \cite[Proposition~4.4]{redondo}):

\begin{theorem}\label{thmhapp}
	If $G$ is a connected  directed graph without oriented cycles and $k$ is an (algebraically closed) field, then 
	\[
	\dim_{k} \mathrm{HH}^i(A)=
	\begin{cases}
		1, \quad &\text{ if } i=0 \\
		0, \quad &\text{ if } i>1 \\
		1-n +\sum_{e\in E(G)}\dim_{k} e_{t(e)}A e_{s(e)},  \quad &\text{ if } i=1
	\end{cases}
	\] 
	where $A=kG$ is the path algebra of $G$,  $n=|V(G)|$ is the number of vertices of $G$ and $e_{t(e)}A e_{s(e)}$ is the subspace of $A$ generated by all the possible paths from $s(e)$ to $t(e)$ in $G$.
\end{theorem}  

A classical result by Gerstenhaber and Schack \cite{GERSTENHABER1983143} gives a topological interpretation of Hochschild cohomology, when restricted to digraphs arising from face posets of simplicial complexes. In particular, if $P$ is the face poset of a (finite) simplicial complex $K$, we have the chain of isomorphisms, for all $i\in \mathbb{N}$,  
\begin{equation}\label{eq:comparison}
\mathrm{RH}_i(G(P))\cong\mathrm{H}_i(K)\cong\mathrm{H}^i(K)\cong \mathrm{HH}^i(kP)
\end{equation}
where $G(P)$ and the first isomorphism  are as in Example~\ref{ex:reachabilityposets}, whereas the last isomorphism with the Hochschild cohomology of the \emph{incidence algebra} $kP$ of $P$ is given by~\cite{GERSTENHABER1983143}.  The central isomorphism between homology and cohomology groups holds because we assume to work with field coefficients. 
We can summarise it as follows:
 
 \begin{proposition}
     Hochschild cohomology of 
     $kP$ and reachability homology of \(G(P)\) are isomorphic, hence they yield the same homology groups and have the same ranks.
 \end{proposition}

Recall that the Betti numbers of a simplicial complex $K$ are the ranks $\beta_i(K)=\mathrm{rk}(\mathrm{H}_i(K))$ of the homology groups of $K$. Analogously, in light of Eq.~\ref{eq:comparison}, we can define the reachability Betti numbers and the Hochschild Betti numbers as the ranks of $\mathrm{RH}_i(G(P))$ and $\mathrm{HH}^i(kG(P))$. 

\subsection{Persistent HH-curves}

Recall that persistent homology can be seen as a functor \((\mathbb{R}, \leq) \to \mathbf{FinVect}\) from the poset of real numbers with values in finite dimensional vector spaces. In the topological setting this can be realised by taking the homology of a filtered simplicial complex. In \cite{persistentHH}, the following persistent Hochschild (co)homology pipeline was introduced, where Hochschild (co)homology is computed  on the condensation $c$ of digraphs:
\begin{equation*}\label{eq:compvar}
    (\mathbb{R}, \leq) \to \mathbf{Digraph}\xrightarrow{c}\mathbf{Digraph}\xrightarrow{k-} k\text{-}\mathbf{Alg}\xrightarrow{\mathrm{HH}}\mathbf{FinVect},
\end{equation*}
that is, at step $n\in (\mathbb{R}, \leq)$, we compute $\mathrm{HH}(k(c(G_n)))$. However, due to the condensation operation, this pipeline is not functorial. In \cite[Definition~5.9]{zbMATH07844814}, the following alternative definition  was proposed:

\begin{definition}\label{def:pHH}
Let $\mathcal{F}\colon (\mathbb{R},\leq ) \to \mathbf{Digraph}$ be a filtration of finite directed graphs. Then, its persistent Hochschild cohomology groups are given by the composition
\begin{equation*}
  (\mathbb{R}, \leq) \xrightarrow{\mathcal{F}} \mathbf{Digraph} \xrightarrow{}  \mathbf{Poset} \xrightarrow{k\text{-}}k\text{-}\mathbf{Alg} \xrightarrow{\mathrm{HH}}\mathbf{FinVect} \ ,
\end{equation*} 
where $\mathbf{Poset}$ is the category of posets, the functor $\mathbf{Digraph} \to \mathbf{Poset} $ sends a digraph $G$ to the poset $\mathcal{R}(G)$, and $\mathrm{HH}$ computes the Hochschild cohomology groups of the algebra $k\mathcal{R}(G)$.  
\end{definition}

For each filtration step $n$ in Definition~\ref{def:pHH}, we obtain a digraph $G_n=\mathcal{F}(n)$, and for $i\in \mathbb{N}$, the $i$-th Betti number 
\[\beta_i^{\mathrm{HH}}(G_n)=\mathrm{rk}(\mathrm{HH}^i(k\mathcal{R}(G_n)))\ .\]
We define the persistent $\mathrm{HH}$-curves as the Betti curves of the persistent Hochschild cohomology groups. Concretely, if $\mathcal{F}\colon (\mathbb{R},\leq)\to \mathbf{Digraph}$ is a filtration of directed graphs, the HH-curves are given by the Betti numbers as a function of the filtration parameter, as the following composition:
\begin{equation*}\label{eq:pers_betti_HH}
  (\mathbb{R}, \leq) \to \mathbf{Digraph} \to \mathbf{Poset} \xrightarrow{k\text{-}}k\text{-}\mathbf{Alg}\xrightarrow{\beta^\mathrm{HH}} \mathbb{N} \ .
\end{equation*}
In view of Equation~\eqref{eq:comparison}, we have that persistent Hochschild cohomology and persistent homology of reachability posets yield the same Betti curves -- cf.~\cite[Proposition~5.10]{zbMATH07844814}:

\begin{proposition}
Let $\mathcal{F}\colon (\mathbb{R},\leq ) \to \mathbf{Digraph}$ be a filtration of  directed graphs. Then, the persistent \(\mathrm{HH}\)-curves agree with the  Betti curves of the order complexes of the  reachability posets.
\end{proposition}

In Section \ref{sec:pipeline} we explain how to use persistent HH-curves and reachability homology in concrete classification tasks.

\subsection{Homology of random digraphs}\label{sec:random_graphs}
In this section we illustrate that the reachability homology of random digraphs behaves very differently from the homology of directed flag complexes. This observation will be essential in the construction of our classification pipeline in Section \ref{sec:pipeline}, specifically in identifying interesting range of filtration values.

Understanding the homological behaviour of random graphs and simplicial complexes has spawned limit theorems and results on the expected Betti numbers of different random models such as the Erdõs-Rényi random graphs~\cite{zbMATH06189102,zbMATH06973992,zbMATH07379056,zbMATH07565231,zbMATH07955456}. Such results are important theoretical foundations for persistence analysis as we need to understand whether the observed homologies of data sets actually deviate from random null models.

\begin{figure}
    \centering
    \includegraphics[
    scale=0.6]{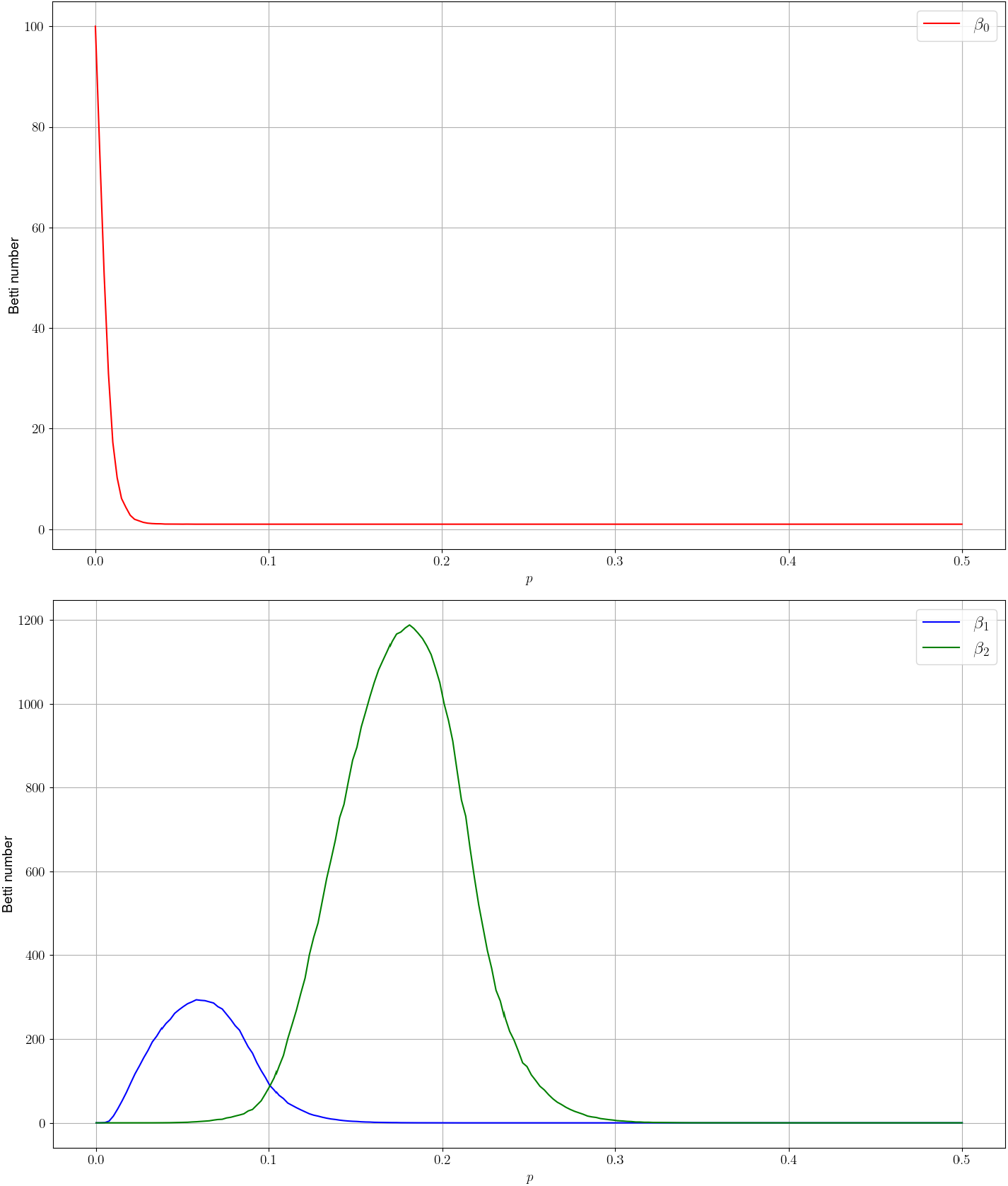}
    \caption{The mean Betti numbers 0, 1, and 2 with respect to the edge probability \(p\) over 200 realisations of the directed flag complex of the Erdõs-Rényi random digraph \(G(100,p)\).}
    \label{fig:raw_ER_bettis}
\end{figure}

To approximate the expected Betti numbers we computed the population mean of a collection of Erdõs-Rényi random digraphs \(G(n,p)\) on \(n=100\) vertices for a range of edge probability values \(p\). For each \(p\), \(r\) realisations of the digraph \(G(n,p)\) were simulated, specified Betti numbers were computed, and the respective mean Betti numbers were taken over the \(r\) realisations. For the Betti numbers of the directed flag complex the range of \(p\) was \([0,0.5]\) divided into 200 equally space intervals; for each \(p\) we took the mean over \(r=200\) realisations. Figure \ref{fig:raw_ER_bettis} shows the mean Betti numbers of the directed flag complexes in degrees 0, 1, and 2.

\begin{figure}
    \centering
    \includegraphics[
    scale=0.3]{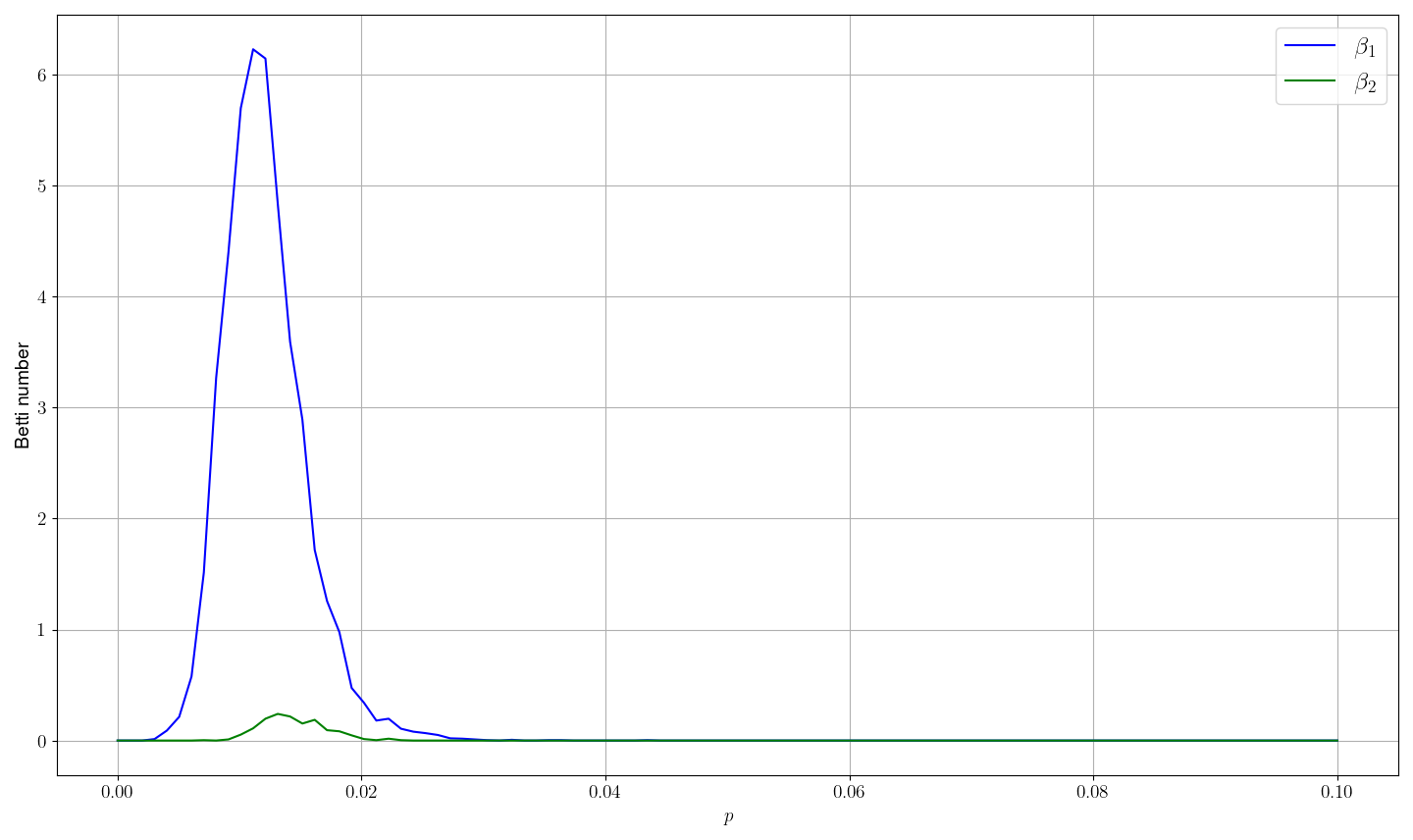}
    \caption{The mean reachability Betti numbers 1 and 2 with respect to the edge probability \(p\) over 300 realisations of the Erdõs-Rényi random digraph \(G(100,p)\). Note that the range of \(p\) is much smaller than in \ref{fig:raw_ER_bettis}, showing that the reachability homology is confined into a very small range.}
    \label{fig:reach_ER_bettis}
\end{figure}

For reachability Betti numbers the range of \(p\) was much smaller, \([0,0.1]\) on 100 intervals. The reduction coming from the condensation operation makes the homology computations much more feasible compared to the case of directed flag complex; hence we took the mean over 300 realisations. The resulting mean Betti numbers are plotted in Figure \ref{fig:reach_ER_bettis}. The figure shows how the non-trivial reachability homology appears in drastically smaller range of \(p\) values than the homology of directed flag complexes; note that the degree 0 homology, i.e.\ connected components, is the same in both cases.

\section{Network classification pipeline}\label{sec:pipeline}

In this section, we describe our method for classifying network data by featurising with persistent Hochschild homology. A generic network data set consists of a collection of edge weighted digraphs and their labels $\mathcal{H} = \{ (H^{\alpha},l_{\alpha})\}_{\alpha \in \Lambda}$ 
, where $l_{\alpha} \in \{0,\dots,k\}$; in this paper we are concerned with binary classification in which \(k=1\). 

\textbf{Thresholding.} The first step of our procedure, and one that has a non-trivial impact on graph classification is thresholding. The idea behind this is that only edges with weights in a certain range are relevant to analysis, so we remove edges below (and/or above) certain threshold(s). As an example, in Section \ref{sec:classification_results} we employ our pipeline on correlation networks where the edge weights represent correlation measures between time series' measured on different brain locations. In such scenarios edges with low or negative values might correspond to unimportant correlations or numerical artifacts. In the pre-processing we want to remove such edges. 
The thresholded graphs \(G^\alpha\) is induced from $H^{\alpha}$ by only those edges whose weights are in a chosen threshold range \([\theta_1,\theta_2]\). The set $\mathcal{G} = \{ G^{\alpha}\}_{\alpha \in \Lambda}$ represents the graphs that we will compute homology features for.







\textbf{Homology features - computing filtration bounds.} As we demonstrated in Section \ref{sec:random_graphs} with random graphs, nontrivial reachability Betti numbers appear only for a small range of edge probability values. We expect that the same might happen for networks arising from a studied data set. Because we do not want feature vectors having a large fraction of zero entries, our next task is hence to identify the range in which the graphs have nonzero Betti numbers. 

The edge weights $\tilde{w}_0 < \tilde{w}_1 < \dots$ of $G$ within the interval \([\theta_1,\theta_2]\) induce a filtration as in Equation~\eqref{eq:graph_filtration}. Let  $G[\tilde{w}_i]$ denote the graph on filtration value $\tilde{w}_i$, that is, $G[\bar{w_i}]$ is induced by edges with weights $\leq \tilde{w}_i$. For $j\neq 0$, we let $k^j(G)$ and $K^j(G)$ be the minimal and maximal weights corresponding to non-trivial \(j\)th homology groups:
\[
k^j(G)=\min \{\tilde{w}_i \mid \mathrm{H}_j(G[\tilde{w}_i])\neq 0\} \text{ and } K^j(G)=\max \{\tilde{w}_i \mid \mathrm{H}_j(G[\tilde{w}_i])\neq 0\},
\]
where the minimum and maximum run across all weights of $G$ in the interval \([\theta_1,\theta_2]\). For the collection of thresholded graphs \(\mathcal{G}\), we set
\[
k^j(\mathcal{G})=\min_{\alpha}k^j(G^\alpha) \text{ and } K^j(\mathcal{G})=\max_{\alpha}K^j(G^\alpha),
\]
which are the global minimum and maximum weight such that all $j$th homology groups of all thresholded graphs are trivial for weights $< k^j(\mathcal{G})$ and $> K^j(\mathcal{G})$. 

Let $n$ be a natural number. For each homology degree $j\in \mathbb{N}$ we subdivide the interval $[k^j(\mathcal{G}),K^j(\mathcal{G})]$ into $n$ subintervals of equal size. Then we consider the points $x_s^j\in [k^j(\mathcal{G}),K^j(\mathcal{G})]$ defined by 
\begin{equation}\label{eq:filtration_steps}
    x_s^j=k^j(\mathcal{G})+s(K^j(\mathcal{G})-k^j(\mathcal{G}))/n
\end{equation}
and use these points, at $s$ varying from $0$ to $n$, as a subsample of edge weights to create a filtration of each graph in \(\mathcal{G}\).

Note that the bounds $k^{j}(G)$ and $K^{j}(G)$ are not universal but depend on the prevalence of non-trivial homology groups, which in turn depend on the used homology theory. Hence, both bounds can be different in our actual pipeline implementations depending on whether the homology computations are done for the directed flag complex or for the reachability poset associated to a digraph \(G\).

\textbf{Homology features - Betti numbers.} For each $j\in \mathbb{N}$, $\alpha\in \Lambda$, and filtration value $x_s^j$ constructed in Equation~\eqref{eq:filtration_steps}, we consider the unweighted directed subgraph $G^{\alpha}[x_s^j]$ of $G^{\alpha}$. Therefore, we get a sequence of subgraphs of $G^{\alpha}$
\begin{equation}\label{eq:filtssampled}
G^{\alpha}[x_0^j]\to G^{\alpha}[x_1^j] \to \dots \to G^{\alpha}[x_n^j]. 
\end{equation}
By computing the \(j\)th homology of the directed flag complex of the graphs in the sequence, we get 
\[
\mathrm{H}_j(\mathrm{dFl}(G^{\alpha}[x_0^j]))\to \mathrm{H}_j(\mathrm{dFl}(G^{\alpha}[x_1^j])) \to \dots \to \mathrm{H}_j(\mathrm{dFl}(G^{\alpha}[x_n^j])) \ ,
\]
hence the Betti curve (in the form of a vector in $\mathbb{N}^{n+1}$)
\[
[\beta_j(G^{\alpha}[x_0^j]),\beta_j(G^{\alpha}[x_1^j]), \dots,\beta_j(G^{\alpha}[x_n^j])] \,
\]
where for ease of notation we set $\beta_j(G^{\alpha}[x_i^j])=\beta_j(\mathrm{dFl}(G^{\alpha}[x_i^j]))$.

For each $\alpha$, we concatenate the Betti curves from homology degrees of interest $j_1 < \dots < j_r$, thus obtaining the vector:
\begin{equation}\label{eq:general_betti_vector}
    B_\alpha=[\beta_{j_1}(G^{\alpha}[x_0^{j_1}]),\dots,\beta_{j_1}(G^{\alpha}[x_n^{j_1}]),\dots, \beta_{j_r}(G^{\alpha}[x_0^{j_r}]),\dots,\beta_{j_r}(G^{\alpha}[x_n^{j_r}])].
\end{equation}

For each $\alpha$, the vector $B_\alpha$ is the feature vector of the graph \(G^\alpha\) corresponding to the filtration of the associated directed flag complex. 

Similarly, we apply the reachability homology to the sequence of graphs in Equation~\eqref{eq:filtssampled}. We get the sequence of reachability homology groups
\[
\mathrm{RH}_j(G^{\alpha}[x_0^j])\to \mathrm{RH}_j(G^{\alpha}[x_1^j]) \to \dots \to \mathrm{RH}_j(G^{\alpha}[x_n^j]) \ ,
\]
from which we get the vector of HH-curves:
\[
B^{\mathrm{HH}}_\alpha=[\beta_{j_1}^{\mathrm{HH}}(G^{\alpha}[x_0^{j_1}]),\dots,\beta_{j_1}^{\mathrm{HH}}(G^{\alpha}[x_n^{j_1}]),\dots, \beta_{j_r}^{\mathrm{HH}}(G^{\alpha}[x_0^{j_r}]),\dots,\beta_{j_r}^{\mathrm{HH}}(G^{\alpha}[x_n^{j_r}])].
\]

\textbf{Homology features - Betti integral.} In our pipeline we also use an approximation of the integral of the Betti curve as a feature vector. For this, let $n$ be the parameter for the filtration subdivision chosen in Eq.~\ref{eq:filtration_steps}. For each $j \in \mathbb{N}$ and $i \in \{1, \dots, n\}$, we consider the 'trapezoidal rule' for approximating the area under the Betti curve between $x_0^j$ and $x_{i}^{j}$ as follows: 
\begin{equation}\label{scurvefeature}
        \gamma_{j}(G^{\alpha}[x^j_i]) = \sum_{k = 0}^{i-1} \frac{(\beta_j(G^{\alpha}[x_{k+1}^j])+ \beta_{j}(G^{\alpha}[x_k^j]))}{2} \cdot (x_{k+1}^{j} - x^{j}_{k}).
\end{equation}
Similarly we have \(\gamma_{i}^{HH}(G^{\alpha}[x^j_i])\) for the Betti curves of reachability homology.


As above, for each $\alpha$ and homology degrees of interest $j_1 < \dots < j_r$ we obtain the feature vectors
\begin{equation}\label{eq:iterated_betti_integral_vector}
    \Gamma_\alpha=[\gamma_{j_1}(G^{\alpha}[x_1^{j_1}]),\dots,\gamma_{j_1}(G^{\alpha}[x_n^{j_1}]),\dots, \gamma_{j_r}(G^{\alpha}[x_1^{j_r}]),\dots,\gamma_{j_r}(G^{\alpha}[x_n^{j_r}])]
\end{equation}
and
\[
\Gamma^{\mathrm{HH}}_\alpha=[\gamma_{j_1}^{\mathrm{HH}}(G^{\alpha}[x_1^{j_1}]),\dots,\gamma_{j_1}^{\mathrm{HH}}(G^{\alpha}[x_n^{j_1}]),\dots, \gamma_{j_r}^{\mathrm{HH}}(G^{\alpha}[x_1^{j_r}]),\dots,\gamma_{j_r}^{\mathrm{HH}}(G^{\alpha}[x_n^{j_r}])].
\]
The element \(i\) of the Betti integral vector is the approximate area under the degree \(j\) Betti curve between \(x_i^j\) and the fixed \(x_0^j\). Hence if the Betti curve is a feature vector of \(n\) elements, the corresponding Betti integral feature vector, at homology degree \(j\), is of length \(n-1\).

\section{Classification results}\label{sec:classification_results}
We implemented the pipeline of Section 2 in Python, using the scikit-learn library. The full pipeline is available in \href{https://github.com/njmead811/Persistent-Hochschild-Homology-Experiments-and-Applications}{Github}. We used support vector machine as the model, implemented using the SVC class with default parameters. We emphasise that this choice of a model, with default parameters, was dictated by our aim of comparing the classification capabilities of reachability homology to that of the simplicial homology of the directed flag complex, and not to achieve best possible accuracy in a specific classification task. 

We used two SVM kernels, linear and RBF. Linear kernel was chosen in order to perform feature ranking to estimate the importance of different homological features. The feature ranking was implemented with the RFECV class, which performs both feature ranking and cross-validation. All feature vectors were standardised with the scikit-learn's StandardScaler class. We implemented our pipeline with all combinations of the choice of a feature vector, SVM kernel, and the homology degrees used in creating feature vectors. Hence, altogether we used 16 different models obtained as combinations from the table below.

\begin{center}
\begin{tabular}{| c | c | c |} 
 \hline
 feature vector  & kernel & Betti numbers\\ 
 \hline  
 $B_{\alpha}$ & &\\
 $B_{\alpha}^{HH}$ & linear with feature ranking & 0, 1 \\ 
 $\Gamma_{\alpha}$ & RBF & 0, 1, 2\\
 $\Gamma_{\alpha}^{HH}$ & &\\
 \hline 
\end{tabular}
\end{center}


\subsection{Epileptic seizure detection}\label{subsec:seizure_detection}
We used the anonymised dataset of 100 recordings of 16 patients in the epilepsy surgery program of the Inselspital Bern \cite{epilepsy_data}. The first 3 minutes of each recording consist of a preictal segment, followed by an ictal segment (between 10s to 2002s) and 3 minutes of postictal time. For us, the first minute of each recording is the baseline data, the 30s before the seizure is the preictal segment and the first 30s of the seizure is the ictal segment. This dataset was also studied in a TDA based machine learning approach in \cite{Caputi2021}. Epilepsy as a very prevalent neurological disorder worldwide, and whose seizures are notoriously difficult to detect from brain activity recordings, has been an active topic for topological analyses \cite{persistence_derivatve, 10.4108/eai.3-12-2015.2262525, piangerelli2018topological, 7163885, stiehl_topological_2023, EEG_visibility_graphs}.

For each recording, let $N$ be the number of scalp electrodes. For every pair of electrode measurements we computed their correlation with the convergent cross mapping (CCM). We then obtain $N \times N$ matrices whose entries are the CCM correlations between the time series associated to each electrode, for the ictal, preictal and baseline segments. These CCM matrices represent the weighted adjacency matrices of digraphs, which are the input to our classification pipeline. We subtract the baseline matrix from both the ictal and the preictal segments to normalise the data; similar normalisation was done in \cite{Caputi2021}. The number of scalp electrodes \(N\) varies between 28-100, depending on the patient, and results in digraphs with different numbers of vertices. Hence homological featurisation is beneficial as it produces global properties not tied to the graph sizes.

Out of the 16 patients only 14 had seizures exceeding 30s. 
For each of these patients we then construct two matrices: one which is obtained by taking the entrywise average of the baseline normalised ictal CCM matrices of that patient and one which is obtained by similarly averaging all of the preictal matrices. For the 14 patients this results in 28 matrices. The classification task is to determine which of the 28 matrices represent preictal or ictal segments.

We use the pipeline outlined in Section 2. There are a variety of hyperparameters, which we summarise below: 

\begin{enumerate}
\item In order to compute the graphs $G^{\alpha}$ in Section \ref{sec:pipeline}, we used different lower threshold values $\theta_1 \in \{ 
-0.4, -0.35, -0.3, -0.25, -0.2, -0.15, -0.1, -0.05\}$. The negative values are due to the normalisation by baseline subtraction. The upper threshold value \(\theta_2\) was chosen to be 0. This was dictated by computational efficiency as it resulted in graph sizes for which the computations were reasonably fast, while not affecting the classification results too much. In addition there were some graphs which had no weights above 0.05.
\item We used $k=2, 3, 5$ folds for cross-validation. Together with the threshold values in item 1., this makes 24 different classification runs we performed.
\item The feature vectors $B_{\alpha}$, $B_{\alpha}^{HH}$, $\Gamma_{\alpha}$, and $\Gamma_{\alpha}^{HH}$ were created using homology degrees $j=0,1,2$. However, for the first half of threshold values none of the graphs had degree 2 reachability homology. Therefore in these cases we created all feature vectors from homology degrees 0 and 1 only. 
\item We subdivided each interval within the filtration bounds into 11 filtration steps by choosing $n = 10$ in Equation~\eqref{eq:filtration_steps}. This choice was made to keep the dimension of the feature vectors reasonably low. In the first half of the threshold values, for each filtration step the Betti numbers \(\beta_0\) and \(\beta_1\) were computed, yielding features vectors $B_{\alpha}$ and $B_{\alpha}^{HH}$ of dimension 22, which is reasonably low compared to the dataset size. In the second half of the threshold values we also used \(\beta_2\) resulting in features vectors $B_{\alpha}$ and $B_{\alpha}^{HH}$ of dimension 33. As explained at the end of Section \ref{sec:pipeline} the feature vectors $\Gamma_{\alpha}$ and $\Gamma_{\alpha}^{HH}$ are of length 20 and 30, respectively.
\end{enumerate}



\subsection{Results}\label{subsec:epilepsy_results}
\begin{figure}
    \includegraphics[width=\linewidth]{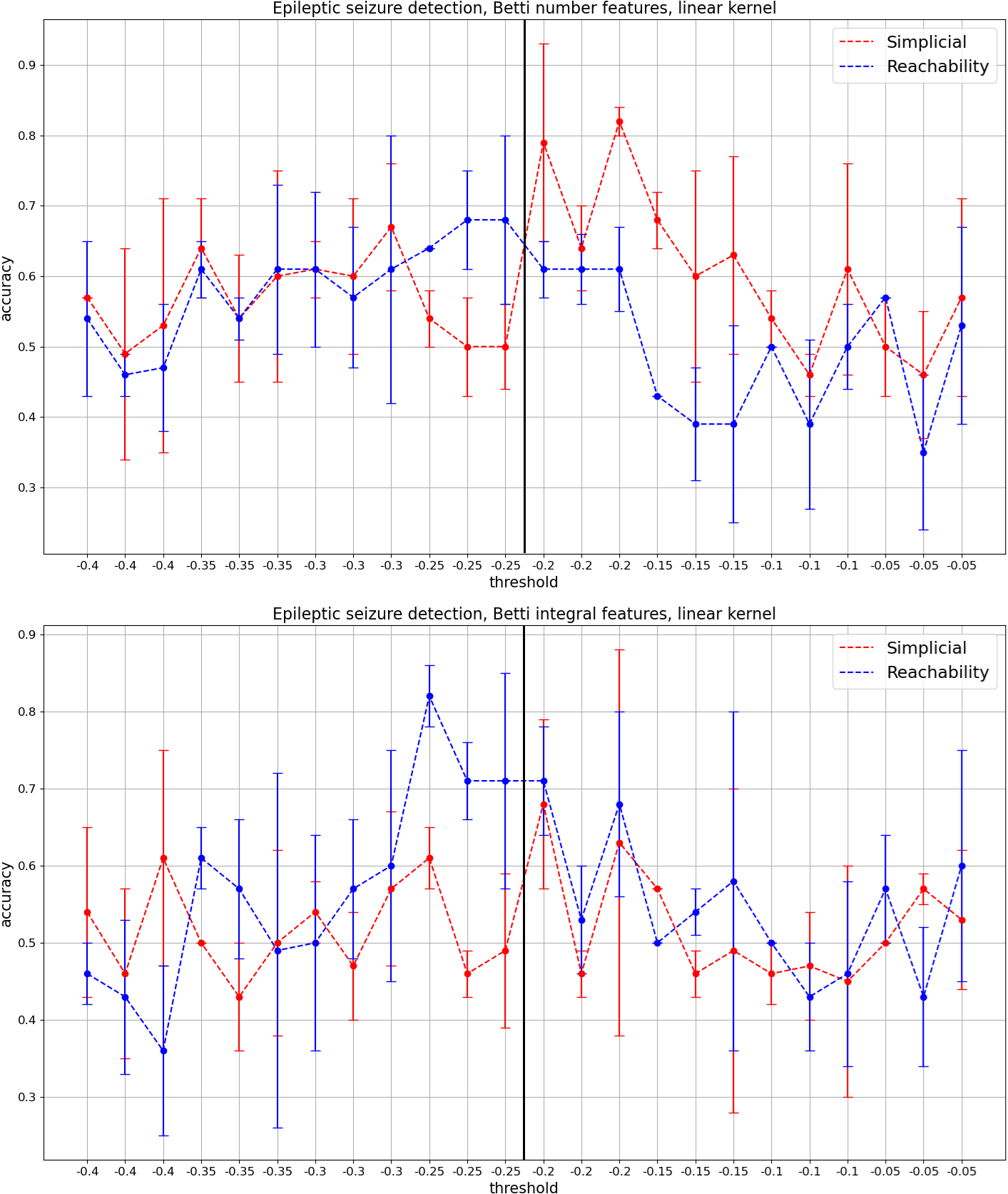}
    \caption{Classification accuracies of SVM with linear kernel in using the Betti number feature vectors (top) and Betti integral feature vectors (bottom). The \(x\)-axis is the lower threshold values used to initially prune the graphs. Each threshold value is replicated three times corresponding to 2-, 3-, and 5-fold crossvalidations, in order. The vertical middle line visually separates the results into left half, where only Betti numbers 0 and 1 are used, and right half where Betti numbers 0, 1, and 2 are used.}
    \label{fig:Epilepsy_linear_accuracies}
\end{figure}

\begin{figure}
    \includegraphics[width=\linewidth]{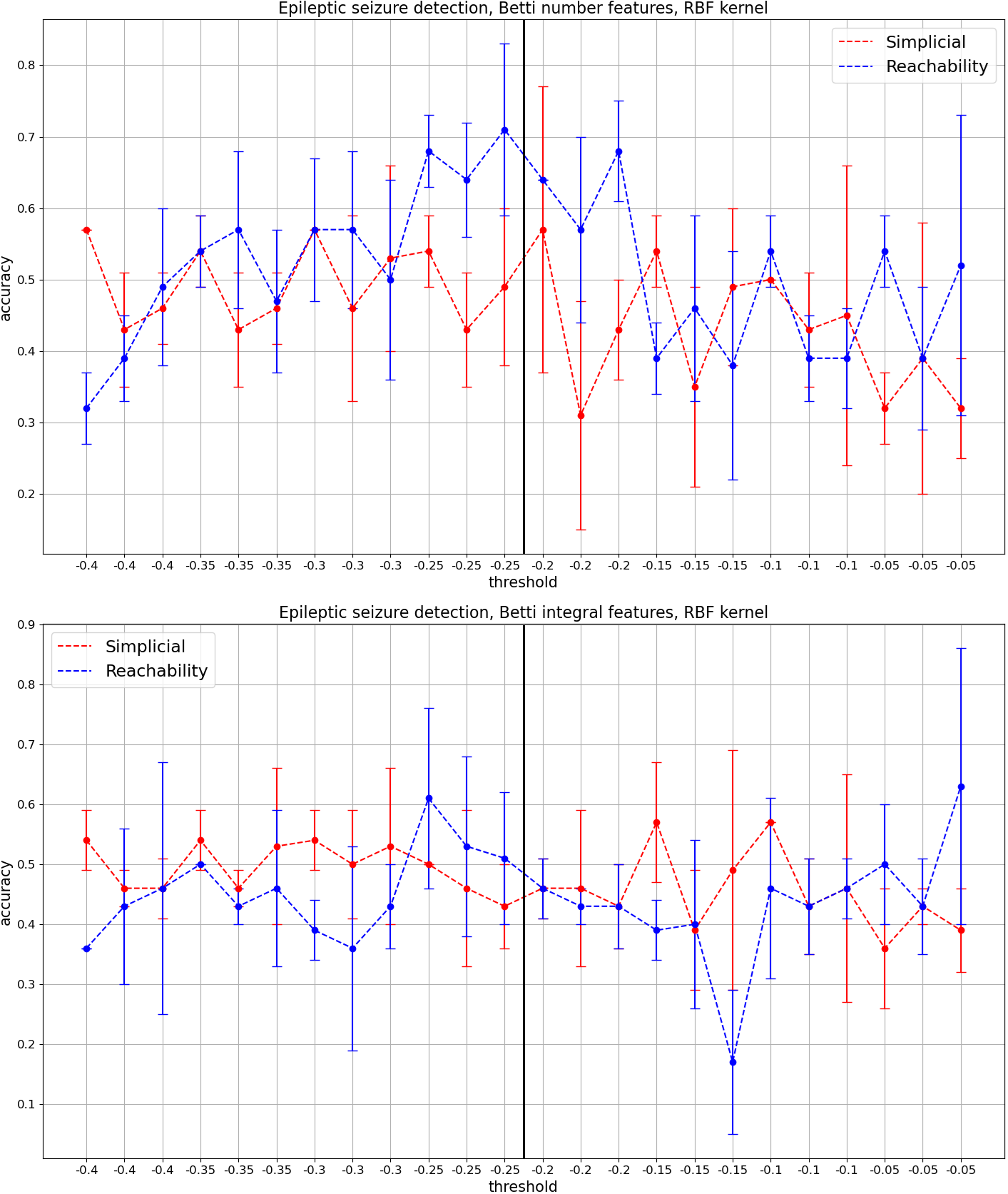}
    \caption{Classification accuracies of SVM with RBF kernel in using the Betti number feature vectors (top) and Betti integral feature vectors (bottom). The \(x\)-axis is the threshold values used to initially prune the graphs. Each threshold value is replicated three times corresponding to 2-, 3-, and 5-fold crossvalidations, in order. The vertical middle line visually separates the results into left half, where only Betti numbers 0 and 1 are used, and right half where Betti numbers 0, 1, and 2 are used.}
    \label{fig:Epilepsy_RBF_accuracies}
\end{figure}

Figures \ref{fig:Epilepsy_linear_accuracies} and \ref{fig:Epilepsy_RBF_accuracies} show the classification accuracies using the linear and RBF kernels, respectively, along with their standard deviations over different thresholds and \(k\)-fold cross-validations for \(k=2,3,5\). We call simplicial homology of the directed flag complex in this section and in the figures just simplicial, similarly reachability homology is simply called reachability. 

In general the best accuracies appear in the mid-range of the threshold values, within -0.3 to -0.2; the exception is the Betti integral features with RBF kernel where the best accuracy appear at the high end of thresholds with reachability homology. In the lower end of thresholds only a fraction of edges are pruned, leaving denser graphs to be homologically featurised; in the upper end the graphs get sparser. In both ends the homological expressivity is reduced in terms of the classification accuracy. 

It is noteworthy that the variability of accuracies for reachability is larger. This is very probably due to the condensation type of operation going from the reachability digraphs to reachability posets, which affects the appearance of homology. Indeed, the digraph below illustrates this point.
\begin{center}
	\begin{tikzpicture}
		\tikzstyle{point}=[circle,thick,draw=black,fill=black,inner sep=0pt,minimum width=2pt,minimum height=2pt]
		\tikzstyle{arc}=[thick,shorten >= 4pt,shorten <= 4pt,->]  

		\coordinate (a3) at (0,0.5);
		\coordinate (b3) at (0.9,1.2);
		\coordinate (c3) at (0.9,-0.2);
        \coordinate (d3) at (1.8,0.5);
		
		\draw[arc] (a3) -- (b3);
		\draw[arc] (b3) -- (c3);
		\draw[arc] (a3) -- (c3);
        \draw[arc] (b3) -- (d3);
        \draw[arc] (c3) -- (d3);
        \draw[arc] (a3) -- (d3);
		
		\node[point] at (a3) {};
		\node[point] at (b3) {};
		\node[point] at (c3) {};
        \node[point] at (d3) {};
	\end{tikzpicture}
\end{center}
Just adding another horizontal edge but in reversed direction makes the whole digraph strongly connected, hence it becomes a single vertex in the reachability poset. This behaviour makes the feature vectors more variable with respect to edge insertion or deletion.

In Table \ref{table:top_accuracies} we summarise the best classification accuracies from the 16 different models tested: four featurisation methods, split between the two SVM kernels and the homology degrees used (left half for 0, 1, right half for 0, 1, and 2). As our aim is to compare simplicial and reachability homology the comparisons of interest are all the adjacent pairs \((B_{\alpha},B_{\alpha}^{HH})\) and \((\Gamma_{\alpha},\Gamma_{\alpha}^{HH})\) compared horizontally in the table. Out of these 8 comparisons, reachability outperforms in 7. Simplicial performs better only with pure Betti number features when using the linear kernel. A possible explanation to this can be found by noting from Figure \ref{fig:Epilepsy_linear_accuracies} that the highest simplicial accuracies are on the threshold -0.2 where \(\beta_2\) is used. By examining the raw feature vectors we observed that in the whole latter half of threshold values reachability has only very few non-zero \(\beta_2\) features, while on the contrary, simplicial has drastically more \(\beta_2\) features, reaching the order of 300-400. Moreover, \(\beta_0\) is always the same for simplicial and reachability and due to the condensation into reachability posets it is expected that reachability also has lower presence of \(\beta_1\). Hence this suggests that simplicial feature vectors can have more distinguishing information in the \((\beta_0,\beta_1,\beta_2)\)-space and the linear SVM model can more easily find the separating hyperplane, explaining the better performance of \(B_\alpha\). From the left half of the top plot in Figure \ref{fig:Epilepsy_linear_accuracies} we see that when using only \(\beta_0\) and \(\beta_1\) features reachability yields the highest accuracy.

\begin{table}[h]
\begin{center}
\begin{tabular}{| c | c | c | c | c ||| c | c | c | c |} 
 \hline
  & $B_{\alpha}$ & $B_{\alpha}^{HH}$ & $\Gamma_{\alpha}$ & $\Gamma_{\alpha}^{HH}$ & $B_{\alpha}$ & $B_{\alpha}^{HH}$ & $\Gamma_{\alpha}$ & $\Gamma_{\alpha}^{HH}$ \\ 
 \hline  
 linear & 67\% & \textbf{68\%} & 61\% & \textbf{82\%} & \textbf{82\%} & 61\% & 68\% & \textbf{71\%} \\
 \hline 
 RBF & 57\% & \textbf{71\%} & 54\% & \textbf{61\%} & 57\% & \textbf{68\%} & 57\% & \textbf{63\%} \\
 \hline 
\end{tabular}
\end{center}
\caption{Best classification accuracies obtained, as split between the different feature vectors and SVM kernels; similarly to Figures \ref{fig:Epilepsy_linear_accuracies} and \ref{fig:Epilepsy_RBF_accuracies} the vertical triple line divides the table by the used Betti numbers: in the left half only \(\beta_0\) and \(\beta_1\) are used, in the right half \(\beta_0\), \(\beta_1\), and \(\beta_2\) are used. The horizontally adjacent pairs of feature vectors \((B_{\alpha},B_{\alpha}^{HH})\) and \((\Gamma_{\alpha},\Gamma_{\alpha}^{HH})\) provide the comparisons between simplicial and reachability homologies. In 7 out the 8 comparisons reachability yields the best accuracy (bolded).}
\label{table:top_accuracies}
\end{table}

\begin{figure}[h!]
    \includegraphics[width=\linewidth]{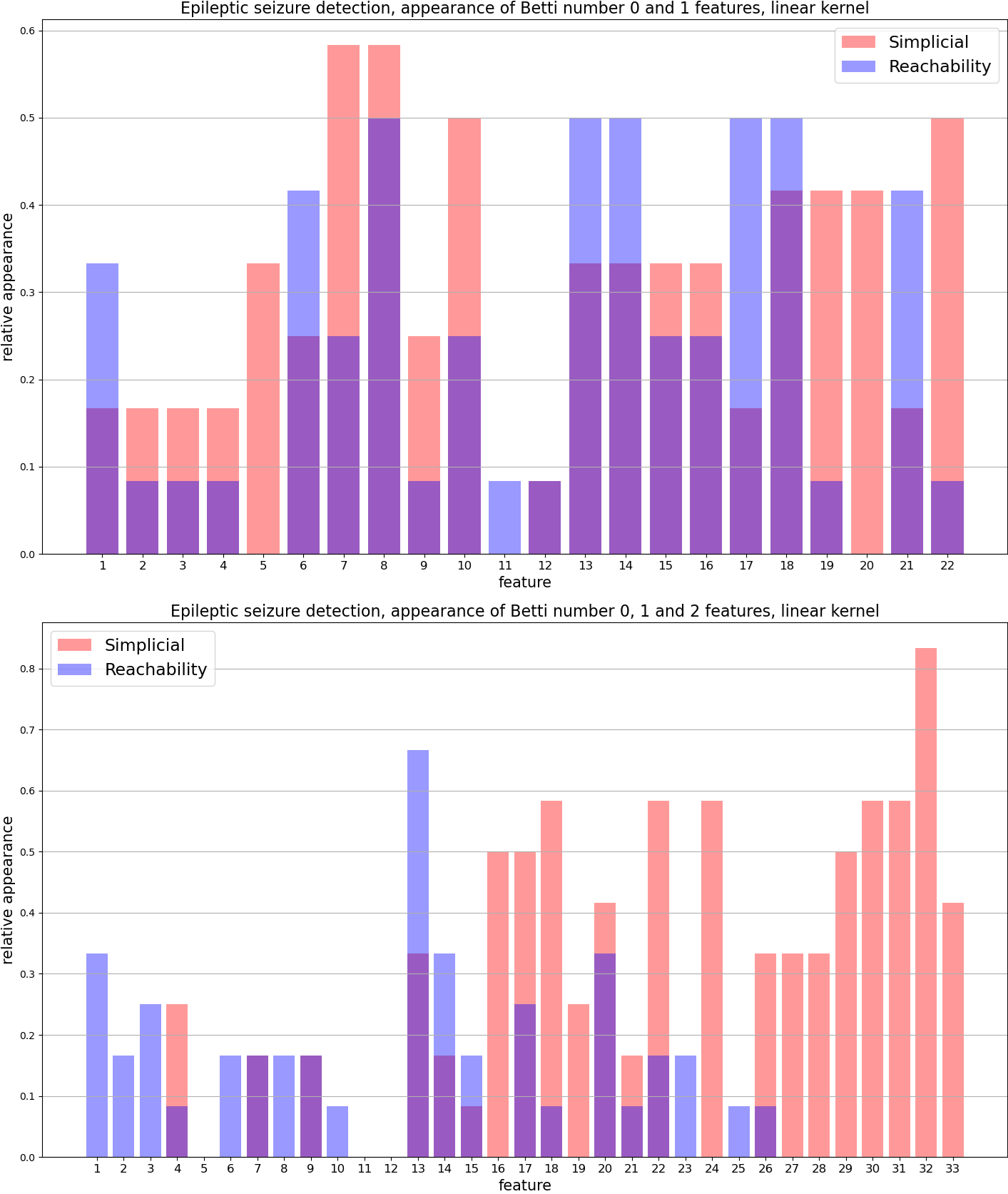}
    \caption{Feature importance in the linear SVM as the appearance frequency of the most important Betti number features, normalised by the number of classification runs. Features 1-11 refer to \(\beta_0\) of the 11 filtration steps in the filtration order, similarly features 12-22 are \(\beta_1\) and features 23-33 are \(\beta_2\).}
    \label{fig:Epilepsy_linear_betti_raw_features}
\end{figure}

Finally, it is of interest to understand how a linear classifier model perceives the importance of homological features. Figure \ref{fig:Epilepsy_linear_betti_raw_features} shows the appearance frequency of different features in Betti number featurisation, as deemed important by  the linear model via the feature ranking in scikit-learn's RFECV class. Each plotted bar is the number of times an actual simplicial or reachability Betti number appers as important for the linear model, normalised by the number of classification runs (24). The top plot shows the feature importance in the first half of thresholds where only Betti numbers 0 and 1 were used, and the bottom plot shows the feature importance in the second half of thresholds where Betti numbers 0, 1, and 2 were used. Features 1-11 refer to \(\beta_0\) of the 11 filtration steps in the filtration order, similarly features 12-22 are \(\beta_1\) and features 23-33 are \(\beta_2\). 

An interesting observation from the top plot is that reachability uses proportionately small number of \(\beta_1\) in the last four filtration steps, albeit the penultimate. This is possibly due to the filtered graphs becoming denser, which makes the condensation collapse larger fractions of the graphs, and hence resulting in vanishing of degree 1 reachability homology. The noticeable difference in \(\beta_2\) is clearly seen in the bottom plot. Simplicial feature importance is dominated by the appearance of \(\beta_1\) and \(\beta_2\) while reachability uses relatively smaller number of \(\beta_1\) and very little to no \(\beta_2\), the latter being due to the essential vanishing of degree 2 reachability homology in all filtration steps. These observations support our possible explanation above, that the better classification accuracy of simplicial with Betti number features and linear kernel is due to the more prevalent higher Betti numbers.

An interesting independent observation from TDA point of view is the proportionately high fraction of \(\beta_2\) features compared to \(\beta_0\) and \(\beta_1\) in the classification via simplicial homology. Often TDA analyses focus on using degree 0 and 1 homologies, largely due to computational efficiency. But our classification problem shows that large part of important distinguishing information resides in degree 2 homology. This points to the question of further understanding the relevance of various homological degrees in different machine learning contexts.

\section{Conclusion}
In this work, we tested the recently introduced persistent reachability homology (PRH) on a network classification task. PRH is related to the (algebraic) Hochschild (co)homology of path algebras of digraphs, and captures different combinatorial information compared to the classical simplicial homology based on directed flag complexes (DPH); in essence, PRH can be seen as a homology theory related to both DPH and Hochschild (co)homology. From a computational viewpoint, PRH captures minimal homological information of digraphs by condensing strongly connected components into single vertices, which makes it significantly faster to compute than DPH.

Our main aim in this work was to investigate the utility of PRH in network classification by comparing its performance to that of DPH. We addressed the prominent classification task of epilepsy detection from EEG correlation networks, by implementing our methodology as an SVM based pipeline. By comparing the classification results of PRH and DPH based network featurisation, we found that out of 8 different cases (different combinations of featurisation method, SVM kernel, and homology degrees used) PRH yielded the best accuracy in 7. Moreover, by using linear SVM kernel for feature ranking, we found that PRH and DPH exploited different sets of features. For instance, the degree 2 Betti numbers were used much more frequently by DPH -- an intriguing independent observation given TDA's typical focus on degrees 0 and 1 for computational efficiency.

Our results point to the need to understand more deeply the generic behaviour of reachability homology with respect to digraph structures, and the relevance of this in TDA practice. In wider scope, our findings demonstrate the value of further investigating the role of various homology theories in TDA applications. Different homology theories can exploit different properties of dataset, such as combinatorial, topological or algebraic features. Thus, their adoption by TDA practitioners can increase the versatility and power of TDA-based machine learning.

\section*{Contributions}

CL: Conceptualisation, Methodology, Investigation, Writing

\noindent NM: Software, Data curation, Formal analysis, Writing

\noindent HR: Conceptualisation, Methodology, Investigation, Writing, Visualisation, Validation, Formal analysis

\bibliographystyle{alpha}
\bibliography{references}
\end{document}